%% file: main.tex
\documentclass{IEEEoj}
\usepackage[noadjust]{cite}
\usepackage{amsmath,amssymb,amsfonts}
\usepackage{algorithmic}
\usepackage{graphicx,color}
\usepackage{textcomp}
\def\BibTeX{{\rm B\kern-.05em{\sc i\kern-.025em b}\kern-.08em
    T\kern-.1667em\lower.7ex\hbox{E}\kern-.125emX}}
\AtBeginDocument{\definecolor{ojcolor}{cmyk}{0.93,0.59,0.15,0.02}}

\usepackage[T1]{fontenc}

\usepackage{subfig}
\usepackage{caption}
\graphicspath{{../pdf/}{../jpeg/}}
\DeclareGraphicsExtensions{.pdf,.jpeg,.png}
\usepackage{amsmath}
\usepackage{amssymb}
\newcommand{\norm}[1]{\left\lVert#1\right\rVert}

\usepackage{array}
\usepackage{color}

\usepackage{makecell}
\usepackage{amsthm}
\usepackage{eqnarray}
\usepackage{hyperref}
\usepackage{subfiles}

\begin{document}
\receiveddate{25 October 2022}
\reviseddate{24 January 2023}
\accepteddate{2 February 2023}
\publisheddate{13 February 2023}
\currentdate{7 March 2023}
\doiinfo{OJSP.2023.3244486}

\title{On The Relationship Between Universal Adversarial Attacks And Sparse Representations}

\author{DANA WEITZNER AND RAJA GIRYES}
\affil{Electrical Engineering, Faculty of Engineering, Tel Aviv University, Tel Aviv 6997801, Israel}
\corresp{CORRESPONDING AUTHOR: Dana Weitzner (e-mail: danaweitzner@mail.tau.ac.il).}
\markboth{On The Relationship Between Universal Adversarial Attacks And Sparse Representations}{Weitzner and Giryes}

\begin{abstract}
  The prominent success of neural networks, mainly in computer vision tasks, is increasingly shadowed by their sensitivity to small, barely perceivable adversarial perturbations in image input. 
  In this work, we aim at explaining this vulnerability through the framework of sparsity. 
 We show the connection between adversarial attacks and sparse representations, with a focus on explaining the universality and transferability of adversarial examples in neural networks.
  To this end, we show that sparse coding algorithms, and the neural network-based learned iterative shrinkage thresholding algorithm (LISTA) among them, suffer from this sensitivity, and that common attacks on neural networks can be expressed as attacks on the sparse representation of the input image. The phenomenon that we observe holds true also when the network is agnostic to the sparse representation and dictionary, and thus can provide a possible explanation for the universality and transferability of adversarial attacks. 
  The code is available at \url{https://github.com/danawr/adversarial_attacks_and_sparse_representations/}.  
\end{abstract}



\maketitle

\section{INTRODUCTION}
\IEEEPARstart{D}{eep} 
neural networks are increasingly used in many real life applications. Therefore, their astonishing sensitivity to small perturbations raises great concerns. Since first reported \cite{adv_examples_introduction}, numerous works have been dedicated to devising defense strategies \cite{defence_1, defence_2, defence_3}, as well as the design of ever more sophisticated attacks \cite{attack_1, attack_2, explaining_1}.
Despite extensive research, considering a wide range of perspectives, it is still unclear why neural networks are so susceptible to these minute perturbations \cite{explaining_1, explaining_2, explaining_3, explaining_4, explaining_5, explaining_6}. 
One of the most intriguing properties of adversarial attacks is the existence of image agnostic (universal) and model agnostic (transferable) adversarial perturbations. It is not fully understood why some adversarial examples generated for one model may hinder the performance of another, and how some perturbations cause the miss-classification of entire datasets. 

When discovered by \cite{universal_adv_attacks}, it was suggested that universal examples occur since there exists some low dimensional subspace of normals to neural networks decision boundaries, which might cause their susceptibility to small adversarial perturbations. 
We are interested in explaining the universality and transferability of adversarial attacks on NN, through the use of sparsity. 
We suspect that this surprising sensitivity and its intriguing properties stem in the data, and are not specific to neural networks models. Natural images are known to reside in a low dimensional subspace, and can be described by a sparse vector over a redundant dictionary matrix. In this work, we hypothesize that adversarial attacks are related to the sparse representation of the input data, which is agnostic to the specific model used, and so it is universal.


Previous works tried to explain the universality of adversarial attacks, using reasoning from a geometric \cite{universal_adv_attacks} or feature-space \cite{explain_universality_3, explain_universality_4} perspectives. \cite{explain_universality_2} noticed that classes tend to be associated with specific directions in the image domain, and connected it to the universality of adversarial perturbation. 
Most of the works regarding transferability are in the context of devising black-box attacks \cite{transferability_3, transferability_4, transferability_6}, with few trying to explain them \cite{transferability_1, transferability_2, transferable}. 
Note that most works focus on explaining either the transferability across different architectures or different images. Only few try to provide a unified view for both \cite{double_universal}.
In our work, we utilize the sparse representation framework to show that common attack directions exist that are correlated with the sparse representation dictionary structure. 

Note that we are not the first to leverage properties of sparse representations to shed light on adversarial attacks. 
\cite{SC_linear_classification_robustness} focus on linear classifiers over sparse codes and provide classification stability guarantees under linear generative model assumptions that are not generally satisfied and are difficult to verify in practice. 
\cite{sulam} provide a theoretical bound on the robust risk of supervised sparse coding, where a linear classifier is learned together with a sparse representation dictionary, relying on the milder existence of a positive gap in the encoded features \cite{encoder_gap}.
In a recent work, Muthukumar and Sulam \cite{Sulam_new} focus on certified robustness and robust generalization of parametric functions composed of a linear predictor and a non-linear representation map. They provide a tighter robustness certificate on the minimal energy of an adversarial example, as well as tighter data-dependent non-uniform bounds on the robust generalization error. They instantiate these results also for feed-forward ReLU networks, however still with a linear final layer. 
Note though that these works cannot be directly applied to common neural network structures as they mainly show the connection to neural networks on simplified models. We take a further step using sparsity and demonstrate its relevance in standard benchmarks involving real practical neural networks. Moreover, we suggest the sparse representation model to explain the universality and transferability of adversarial examples. 

In this work, we show the connection between universal adversarial attacks to sparse representations, by going in both directions; from neural networks to sparsity and back. 
Specifically, we show that adversarial attacks of neural networks are observable in the sparsity framework and that modifications of the sparse code of input data can be adversarial to neural networks.
This is of interest since many of the phenomena of adversarial perturbation have simple explanations considering a sparse representation of the data. 
In the framework of sparse representations, the success and robustness of algorithms are strongly influenced by the properties of the underlying data dictionary. This can be seen as a reason for the transferability of attacks between different methods, as the sensitivity stems from the dictionary of the data.
The universality among different images stems in the shared dictionary, where the addition of adversarial perturbation that is not sparse, can hurt the representation of many (or all) data points by exploiting specific vulnerabilities in the representing dictionary. This may explain some common practices for improving the robustness of neural networks, such as feature sampling and denoising \cite{activation_pruning}, and promoting the orthogonality of the learned weights \cite{parseval}. The connections we propose suggest that these techniques can also be explained under a sparse representation model, where a valid representation is sparse, and low mutual coherence of the dictionary is a key factor in performance guarantees. \looseness=-1

\textbf{Contribution.}
To show the relationship between sparsity and universal attacks, we perform the following steps. 
First, we describe the direction from neural networks to sparsity in Section \ref{sec:transf_sparse_coding}. We show that the transferability of adversarial attacks extends to the realm of sparsity. Through a series of experiments, on synthetic and real data, we examine the sparsity properties of adversarial perturbations obtained by standard methods on neural networks. Not only do we show the connection in the task of sparse coding, but also in popular classification networks operating on real data, which we take as an example of general neural networks. 

Then, we explore the opposite direction, i.e., from sparsity to neural networks in Section \ref{sec:sparsity_to_nn}. We present a sparsity based attack that we refer to as "Dict-Attack" (DA), which exploits sensitive directions in the representation dictionary. We show that it causes excess error even in general neural networks. While its attack success rate is not as the state-of-the-art strategies, the goal of this approach is to stress the relationship between sparsity and the neural network vulnerability to universal adversarial attacks.  
As it is based on the dictionary representing of the data, our attack is oblivious to a specific model or data point, demonstrating the universality and transferability of adversarial attacks and connecting it to the sparse representation of the input.


\textbf{Outline.} The paper is organized as follows. We show the transferability of adversarial attacks in sparse coding in Section \ref{sec:transf_sparse_coding}, where we examine the adversarial robustness of sparse coding algorithms. In  Section \ref{sec:SC_analysis_of_cls_pgd}, we show how attacks of general neural networks affect the sparse representation of the data. We then develop the other direction and show how changes in the sparse code of the input hurt the performance of general neural networks in Section \ref{sec:sparsity_to_nn}. Finally, we conclude in Section \ref{sec:conclusion}. Full implementation details are specified in the Appendix.

\section{ADVERSARIAL ATTACKS TRANSFERABILITY IN SPARSE CODING}\label{sec:transf_sparse_coding}

\begin{figure*}
\centering
    \subfloat[OMP]{\includegraphics[width=0.32\linewidth]{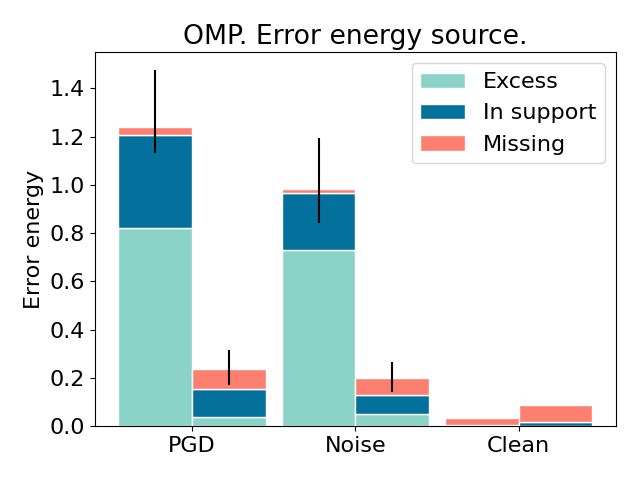}
    \label{fig:attacks_on_omp}}
    \hfil
    \subfloat[LASSO]{\includegraphics[width=0.32\linewidth]{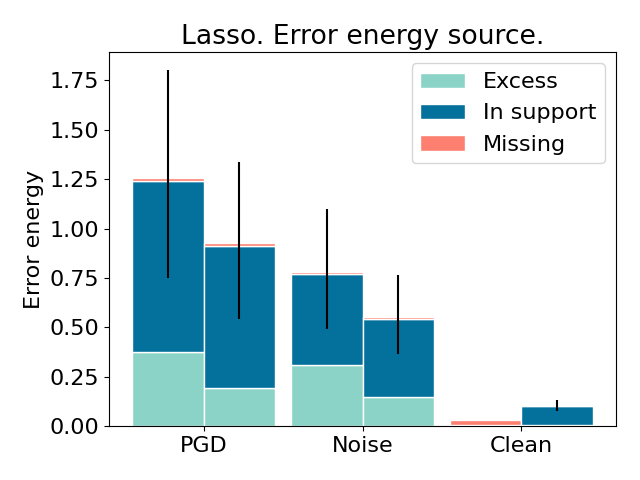}
    \label{fig:attacks_on_lasso}}
    \hfil
    \subfloat[LISTA]{\includegraphics[width=0.32\linewidth]{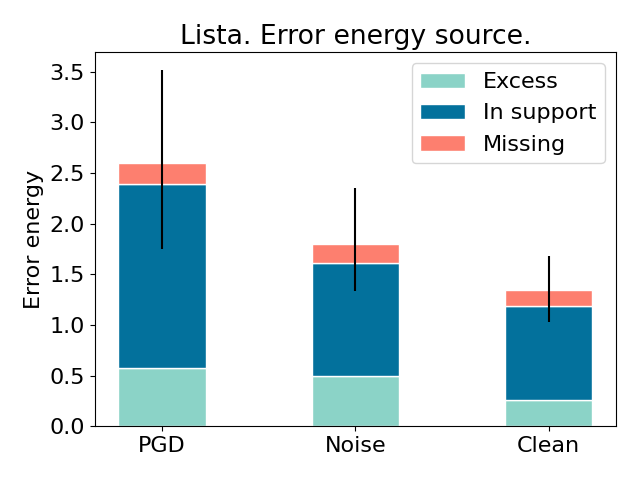}
    \label{fig:attacks_on_lista}}
\caption{
    {\bf Adversarial perturbations on sparse coding algorithms.} 
    We compare PGD to added random noise on three sparse coding algorithms - OMP, LASSO, and LISTA. The attack is calculated on LISTA and transferred to the other methods. In OMP and LASSO, the two bars in each attack stand for two operating modes. OMP with a predefined output sparsity (right), or with a tolerance (left). LASSO with a high (right) or low $\ell_1$ coefficient. 
    The bars show the mean error of the output sparse codes, with colors referring to the original support: "excess" for redundant atoms, "missing", and the error of retrieved components in the correct support ("in support") (see Section \ref{sec:transf_sparse_coding}). The black lines show the 0.25 and 0.75 distribution percentiles. PGD is stronger than random noise not only on its originating LISTA but also on OMP and LASSO.} 
\label{fig:attacks_on_SC_algorithms}
\end{figure*}
If the neural network sensitivity to adversarial perturbations is related to the sparse representation of the data, then we must see the effect of adversarial attacks on this sparse representation. To check the validity of this point, we empirically analyze a common adversarial attack method, namely, the projected gradient descent (PGD) attack \cite{defence_3} in the context of sparse representation. Specifically, we begin with the task of sparse coding, i.e., the retrieval of the sparse representation of a given input, and then move to classification tasks.


We turn to show that the peculiar sensitivity of neural networks is not endemic to them, and show that other, classical, sparse coding methods are susceptible to the same adversarial perturbations.
We start by showing that general sparse coding algorithms, that are not implemented using a neural network, are also sensitive to the same adversarial perturbations, which are generated for neural network based sparse coding. To see this, we utilize the learned iterative shrinkage thresholding algorithm (LISTA, \cite{Lista}), which is a neural network that performs sparse coding. As it is a valid neural network, we can use standard methods to attack it. We focus on the projected gradient descent (PGD) attack \cite{defence_3}, a simple but sufficient method, applied to maximize the reconstruction loss (MSE). Thus, the perturbation is calculated via an optimization process that maximizes the loss such that the perturbation norm is bounded by $\epsilon$. 

Consider the case of sparse data, induced by a dictionary $D \in \mathbb{R}^{m \times n}$ with normalized columns. Thus, each data point $x \in \mathbb{R}^{m}$ is sparse under the dictionary $D$, i.e., $x = D \alpha$,
where $\alpha \in \mathbb{R}^{n}$ is $s-$sparse. We train LISTA using $\ell_2$ loss and batches of data pairs $\{x, \alpha\}$, generated using a normalized Gaussian dictionary. The non-zero entries of $\alpha$ are i.i.d standard Gaussian, in locations sampled uniformly at random. We attack LISTA via a standard $\ell_2$ PGD attack, with $\epsilon = 0.3$ and 40 iterations. In this case, the choice of $\epsilon$ is arbitrary, where it is chosen to be much smaller than the $\ell_2$-norm of the vector $x$. We repeated the experiments with varying values of epsilon and got very similar results. For more details, please see the supplementary material.
We train Lista for 1000 epochs of 200K examples, randomly generated on the fly, simulating infinite data.
To witness the effect of the attack, we compare the attacked input's representation to that of added random noise with the same $\ell_2$ norm as the adversarial perturbation (similar results for $\ell_\infty$ norm are omitted for brevity).

Figure \ref{fig:attacks_on_lista} shows the PGD attack results on LISTA on $1000$ test signals (the same in all experiments). We report the squared error of the representation reconstruction with its 0.25 and 0.75 distribution percentiles. As expected, PGD is statistically significantly stronger than random noise (with a very low p-value in a standard t-test). We divide the error energy bar into fractions that show on what entries the error occurred. Thus, we show the error of excess components (that were zero in the original input), missing components (that were not zero in the original representation and are zero after the attack), and the error of the coefficients in the correct support of the sparse representation. 

We test the transferability of the perturbation calculated with PGD for LISTA to other sparse coding techniques.
The first method is orthogonal matching pursuit (OMP) \cite{omp}, an iterative algorithm that greedily finds the dictionary columns that are most aligned with the input data, given the dictionary $D$. It can be employed using a predefined sparsity level, where the algorithm stops after finding the most contributing $s$ atoms, or by defining a desired error between the input data and its sparse reconstruction, $\hat{x} = D \alpha$. In the latter, the solution does not have a pre-defined sparsity, and OMP is more sensitive to the transferred attack than to added random noise, as can be seen in Figure \ref{fig:attacks_on_omp}. Note though that when using a pre-defined sparsity, the error is much lower as many of the excess components in the attack are being truncated. Statistically, the error distributions cannot be distinguished (0.58 p-value in a t-test), i.e., PGD is no stronger than random noise. This can be seen as the sparse coding equivalent to the feature pruning that is practically performed to enhance the robustness of neural networks \cite{activation_pruning}.


We repeat the same experiment with another sparse coding method, LASSO, where we find the sparse code by solving
\begin{equation}
    \alpha \in \underset{\alpha}{\arg\min}  \frac{1}{2} \norm{x - D \alpha}_2^2 + \beta \norm{\alpha}_1,
\end{equation}
where $\beta$ is a balance hyper-parameter that affects the sparsity of the minimizer. 
In Figure~\ref{fig:attacks_on_lasso}, it is apparent that the PGD attack, obtained from the gradients of the LISTA network, also sabotages LASSO more than noise (again, with a very low p-value in a t-test). We repeat the experiment with a lower $\beta$, which promotes less sparse solutions (the left bars of LASSO in Figure \ref{fig:attacks_on_lasso}). There, we see, similarly to OMP, lower robustness to adversarial and random perturbations alike, and also a greater sensitivity to PGD compared to random noise. This again may explain the success of feature pruning and denoising in improving robustness as larger $\beta$ in LASSO implies more denoising and pruning.


\begin{figure}
\centering
    \subfloat[PGD]{\includegraphics[width=0.5\linewidth]{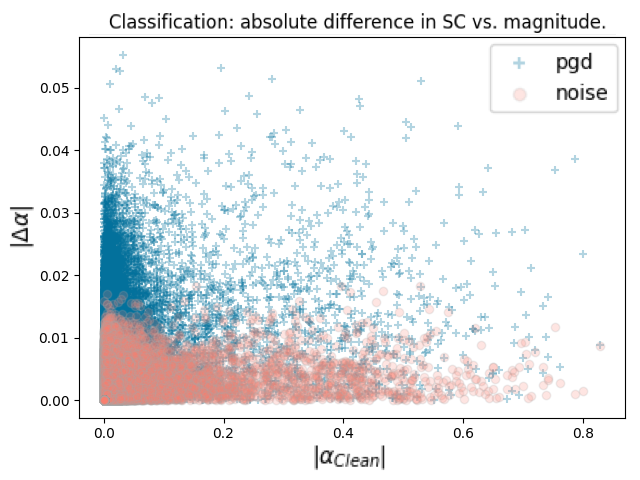}}
    \hfil
    \subfloat[AutoAttack]{\includegraphics[width=0.5\linewidth]{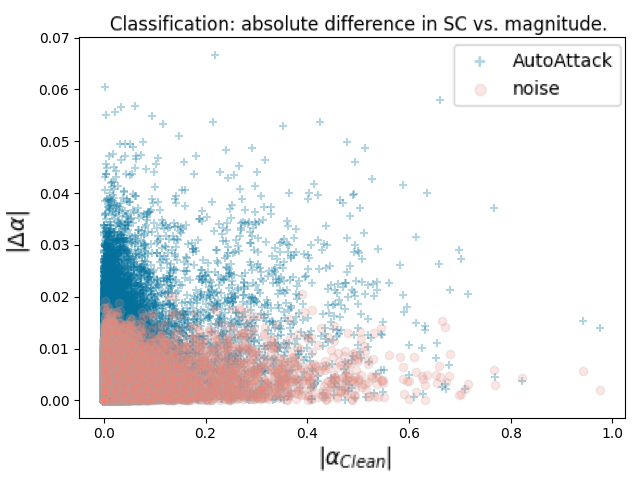}}
\caption{
    Sparse code analysis of PGD (left) and AutoAttack (right) on WideResNet28, compared to added random noise, on a single example from CIFAR-10. Each point represents an entry in the sparse code. 
    We attack WideResNet28 with PGD or AutoAttck and produce the sparse code of the attacked image using ConvLISTA. In addition, we produce the code of the original image and the code of the original image with added noise. 
    The x-axis is the size of the corresponding coefficient in the original ("clean") code.
    The y-axis is the absolute difference between the coefficient of the perturbed image (i.e., attacked or with added noise) and the coefficient of the original image. 
    While the noise adds a somewhat bounded difference, the attacks obtain much higher disruptions, which are even more prominent when the original component is small.}
\label{fig:classification_SC_support_analysis}
\end{figure}

\section{SPARSE CODING IS SENSITIVE TO ADVERSARIAL ATTACKS IN CLASSIFICATION}\label{sec:SC_analysis_of_cls_pgd}
The former analysis was held in sparse coding. Now we turn to examine adversarial attacks on neural networks in the more common task of classification. To this end, we conduct the following experiment, using CIFAR-10 and WideResNet28, which is a standard baseline to test adversarial robustness \cite{adversarial_robustness_benchmark, pang_2022_wide_resnet, gowal_2021_wide_resnet}. We attack a trained WideResNet28 using a standard PGD ($\ell_\infty$ norm, $\epsilon = 8 / 255$, 40 iterations) and also using AutoAttack \cite{AutoAttack}, a combination of several common attacks (APGD, APGDT, FAB, and Square attack). 
Unlike the synthetic case, here we do not have a predefined dictionary representing the data. To be able to see the effect of the attack on the sparse code, we learn a dictionary for CIFAR-10, using ConvLista \cite{Raja_conv_dict}. This model trains LISTA together with a convolutional dictionary (For full implementation and training details, please see the supplementary material). 
We then calculate the sparse representation under this dictionary of inputs with both randomly added noise and adversarially added perturbations of the same magnitude. Notice, that the two networks, i.e., the classifier and the sparse coder, are completely oblivious to one another and trained to perform totally different tasks. As many other computer vision models, both networks are convolutional. We leave the inspection of the architecture effect on the transferability of the attacks to future work.

Since in this case, the data is not-synthetic, the achieved representation is not exactly sparse. Thus, the previous division of the representation MSE w.r.t the support performed in Section~\ref{sec:transf_sparse_coding} is no longer relevant. Therefore, in Figure~\ref{fig:classification_SC_support_analysis}, we scatter plot the magnitude of the difference in the representation (i.e, before and after the attack) vs. the magnitude of the clean code components, for a random sample in CIFAR-10. Each point represents one component in the sparse representation. This is a continuous presentation of the MSE division performed in Section~\ref{sec:transf_sparse_coding}, where the "excess" components are those that were almost zero in the clean representation. Note that, indeed, the small perturbation in input space has a dramatic effect on the sparse representation, even though the classifier and the sparse coder have no direct connection. While the noise adds a somewhat bounded difference, the attacks (PGD and AutoAttack) obtain much higher disruptions, which are even more prominent when the original support is small. This shows that the attacks significantly change the support of the input for fooling the network (with a very low p-value in a standard t-test). Moreover, we see that the type of attack barely affects this phenomenon, thus we conduct the rest of the experiments with PGD solely. 

\begin{figure*}
\centering
    \subfloat[Correlation density histogram]{\includegraphics[width=0.25\linewidth]{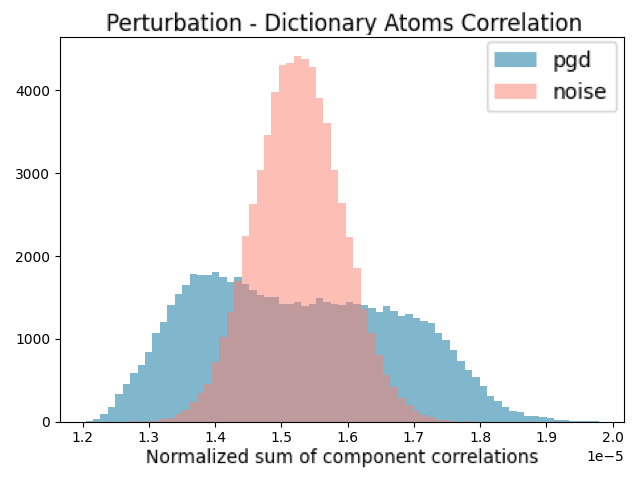}\label{fig:PGD_dict_corr_l2}}
    \hfil
    \subfloat[Sub matrices spectra]{\includegraphics[width=0.25\linewidth]{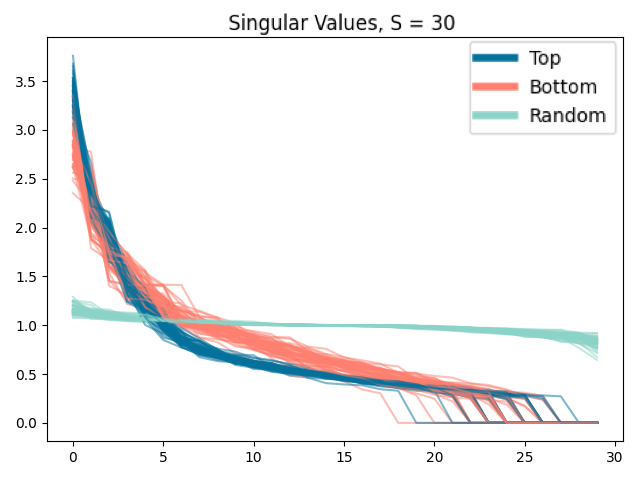}\label{fig:l2_data_spectra}}
    \hfil
    \subfloat[Correlation density histogram]
    {\includegraphics[width=0.25\linewidth]{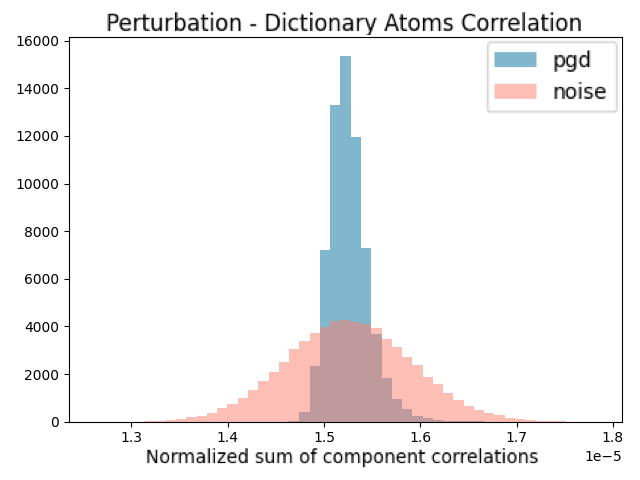}\label{fig:PGD_random_dict_corr_l2}}
    \hfil
    \subfloat[Sub matrices spectra]{\includegraphics[width=0.25\linewidth]{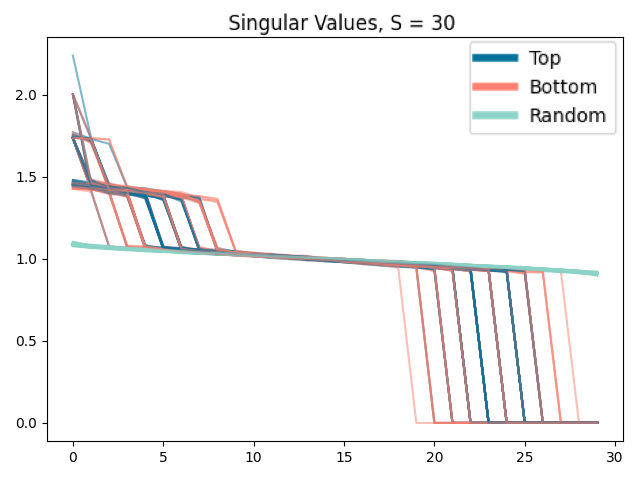}\label{fig:l2_random_spectra}}
\caption{\ref{fig:PGD_dict_corr_l2}: PGD perturbations tend to be more correlated / orthogonal w.r.t dictionary atoms ($\ell_2$). \ref{fig:l2_data_spectra}: The singular values of sampled \textbf{data} dictionary sub-matrices, comprised of the atoms that are highly correlated to PGD perturbations ("top"), orthogonal to them ("bottom"), and random sub-matrices. PGD picks the worst conditioned sub-matrices in the dictionary. \ref{fig:PGD_random_dict_corr_l2}, \ref{fig:l2_random_spectra}: The same only with a \textbf{random} dictionary.} 
\end{figure*}\label{fig:pgd_correlations}

For a wider view of the dataset, we analyze the correlation between PGD perturbations and the dictionary atoms. This is a generalization of the former single example case (shown in Figure \ref{fig:classification_SC_support_analysis}) as it shows the correlation over all the data. 
We start by testing the correlation between adversarial perturbations created by PGD and the dictionary representing the data. Our goal is to demonstrate that the attack hurts the network by highly altering the sparse representation of the input. In the non-synthetic setup (CIFAR-10), we attack WideResNet28 with PGD (using $\ell_2$ with $\epsilon = 128/255$, omitted for brevity for $\ell_\infty$). We then calculate the correlations between the PGD attack perturbations and the normalized atoms of the dictionary representing the data, which was separately learned using ConvLISTA \cite{Raja_conv_dict}. We sum the magnitude of the correlations per atom through the default test set of CIFAR-10, and finally, normalize by the total sum of the correlations (of all atoms). 
Thus, 
\begin{equation}
    \text{corr}_i = \frac{\sum_{j \in [N]} \delta_j^T D_i}{\sum_{j \in [N], i \in [N_D]} \delta_j^T D_i},
\end{equation}
where $N$ is the number of images in the dataset, $N_D$ is the number of atoms in the dictionary, and $\delta$ is the normalized perturbation.
This can be seen as the probability of the perturbation to "pick" a certain dictionary atom. To examine the effect of the adversarial perturbation, we compare it to random noise of the same norm. The $\epsilon$ values we use are standard in adversarial robustness benchmarks on CIFAR10 and MNIST (e.g., \cite{adversarial_robustness_benchmark}).

Figure \ref{fig:PGD_dict_corr_l2} shows the results of the $\ell_2$ attacks. There, we show the distribution of the correlation density of all dictionary atoms. The PGD perturbations tend towards stronger, or weaker correlations with dictionary atoms in average, w.r.t noise. For random noise, the densities seem to follow a Gaussian distribution, which makes sense as the sum of uniform correlations. The PGD perturbations, however, have more extreme probabilities (statistically significant, with a very low p-value in a two-sample Levene test for equal variances). They tend to be either more aligned with certain dictionary atoms or less correlated to other atoms, i.e., more orthogonal to those atoms. This motivates the use of orthogonal regularization of learned dictionaries. However, adding such regularization usually comes at the price of reconstruction accuracy; We leave this for future research.

Since the PGD perturbations pick specific directions, we are interested in the dictionary sub-matrices that are comprised of atoms that are highly correlated or orthogonal to them. In Figure \ref{fig:l2_data_spectra} we plot the singular values of such sub-matrices, that were randomly sampled in the following manner. We sample 50 sub-matrices in each of three categories: atoms that are highly correlated to PGD attacks ("top"), more orthogonal to them ("bottom"), and a test group of random dictionary atoms ("random"). For the top and bottom groups, we sample $30$ out of the $60$ most extreme atoms. For the random groups, we sample $30$ atoms from the entire dictionary. 
Random sub-matrices tend to be orthogonal, and hence their singular values tend to be about one. We suspect that the bottom matrices are generally bad sub-matrices in the dictionary, regardless of the PGD directions. 
PGD, however, manages to select the worst directions in the data, even worse than the naturally occurring sub-matrices. This is apparent in the conditioning of the top sub-matrices, which is worse than the bottom ones. Whether looking at the distributions of the correlation, or in the conditioning of the sub-matrices, it appears that $\ell_2$ PGD attacks are stronger w.r.t the dictionary representing the data. 

To show that the PGD perturbations are indeed correlated with directions in the input data, and not only reside in general random subspaces, we repeat the above experiment in the same setup, only using a random dictionary. Thus, we calculate the normalized correlations between PGD attack perturbations and atoms of a random Gaussian dictionary (which is of the same size as the data driven dictionary). To examine the effect of the adversarial perturbation, we compare it to random noise of the same norm. 
Figure \ref{fig:PGD_random_dict_corr_l2} shows that the PGD perturbations have similar correlations with most atoms, implying that they are of no special direction w.r.t. the random dictionary. In addition, the narrower distribution indicates that the PGD perturbations are in a smaller set of directions with respect to Gaussian noise. This makes sense as the perturbations are calculated on a trained classifier, fitted to the small subspace of natural images. Figure \ref{fig:l2_random_spectra} plots the singular values of the random dictionary sub-matrices, divided by their correlation with PGD perturbations. Notice that also when examining the singular values, the random dictionary does not capture the difference in the correlations of the different PGD attacks. 
This experiment further emphasizes the correlation between the directions selected by PGD and the \emph{learned} dictionary, where there the perturbations have a tendency to a certain subset of the atoms in the dictionary. 
This might be interpreted as a motivation to promote orthogonal weights in neural networks, which limit the potential of the adversary to hinder many representation coefficients given an attack budget.
\begin{figure}[!h]
\centering
    \includegraphics[width=0.6\linewidth]{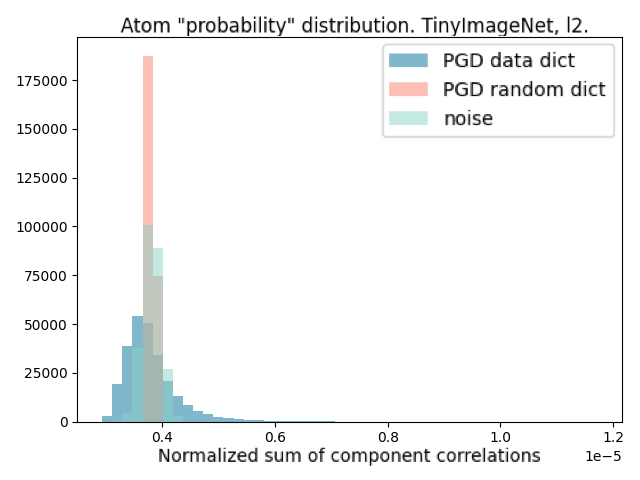}\label{fig:tiny_imagenet_PGD_dict_corr_l2_hist}
\caption{Also on Tiny ImageNet, PGD perturbations tend to be more correlated / orthogonal w.r.t dictionary atoms ($\ell_2$). The tail of PGD with the data dictionary distribution hits $1.2$ units, whereas the noise is of about $0.45$. }
\label{fig:pgd_correlations_tiny_imagenet}
\end{figure}

To further validate our result, we repeated the experiment with another dataset, the more realistic Tiny ImageNet \cite{tiny_imagenet}, with Resnet50. Figure \ref{fig:pgd_correlations_tiny_imagenet} shows the correlation density distribution, which is similar to the CIFAR-10 case.
\section{NEURAL NETWORKS ARE SENSITIVE TO CHANGES IN THE SPARSE CODE OF THEIR INPUTS}\label{sec:sparsity_to_nn}
To show the sensitivity of neural networks to changes in the sparse representations of their input, we devise an attack that exploits sensitive directions in the representation dictionary. We do not aim at beating state-of-the-art methods; The goal of this attack is strictly to show the connection between the two worlds. We show that this universal attack, shared among all input data, as it is based on the dictionary that represents them all, damages the performance of neural networks.

\subsection{THE REPRESENTATION OF AN ADVERSARIAL PERTURBATION IN A DICTIONARY}

To motivate an adversarial attack that relies on the representation dictionary, we analyze the sparse code of the adversarial perturbation itself, detached from its originating input. To this end, we take the adversarial perturbation and produce its sparse code using LASSO. We start by analyzing the representation of PGD attacks in the synthetic setup described in Section~\ref{sec:transf_sparse_coding}. We attack the trained LISTA and then use LASSO to find the representation of $\delta_{PGD} = x_{PGD} - x$ under the learning dictionary of LISTA. To measure the effect of the attack, we compare this sparse code to the sparse code of random noise with the same $\ell_2$ norm.

\begin{figure}[h]
\centering
    \subfloat[Non zero components distribution]{\includegraphics[width=0.45\linewidth]{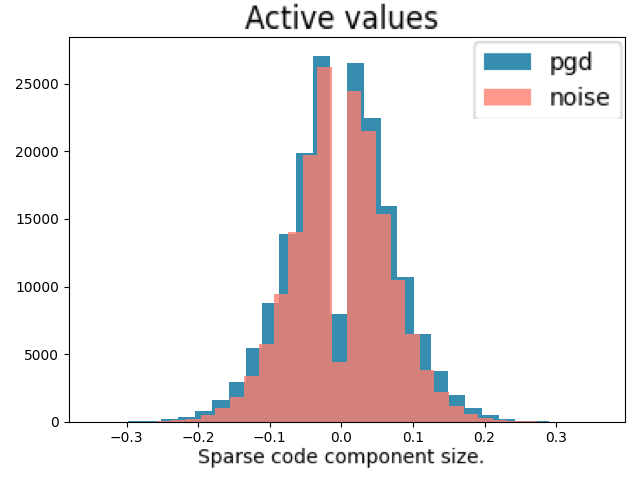}
    \label{fig:delta_hist}}
    \hfil
    \subfloat[Representation energy]{\includegraphics[width=0.45\linewidth]{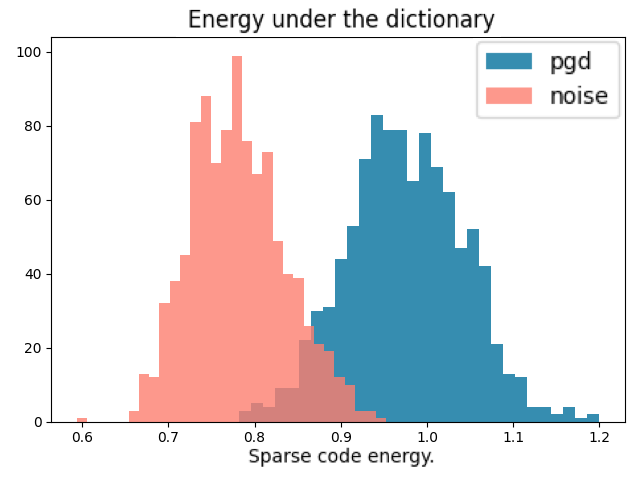}
    \label{fig:delta_energy_hist}}
\caption{
    Sparse code analysis of PGD perturbations compared to random noise, under the dictionary representing the data (synthetic). We attacked LISTA using PGD, and used LASSO to produce the sparse code of the perturbation, i.e., the attacked image minus the original one. On the left, the distribution of non zero components in the sparse codes of PGD perturbation (Blue) and noise (Orange). On the right, the energy distribution of a thousand test samples. PGD has larger and more abundant active components under the dictionary, which results in higher mean energy ($0.976$ of PGD vs. $0.782$ for random noise). } 
\label{fig:delta_analysis}
\end{figure}

Figure \ref{fig:delta_hist} shows the histograms of the non zero components in both sparse codes for 1000 signals. The PGD representation components tend to be both more abundant and larger. This confirms that the PGD perturbation is not sparse under the dictionary (more than noise). Examining the energy (sum of squares) distribution of the codes (Figure \ref{fig:delta_energy_hist}) gets us to the conclusion that PGD perturbation have significantly higher energy under the dictionary (with a very low p-value in a standard t-test). 
This motivates us to design the following attack, which directly finds the perturbation vector with the highest energy under the dictionary, given a predefined norm in the input domain.

\begin{figure*}
\centering
    \subfloat[OMP]{\includegraphics[width=0.3\linewidth]{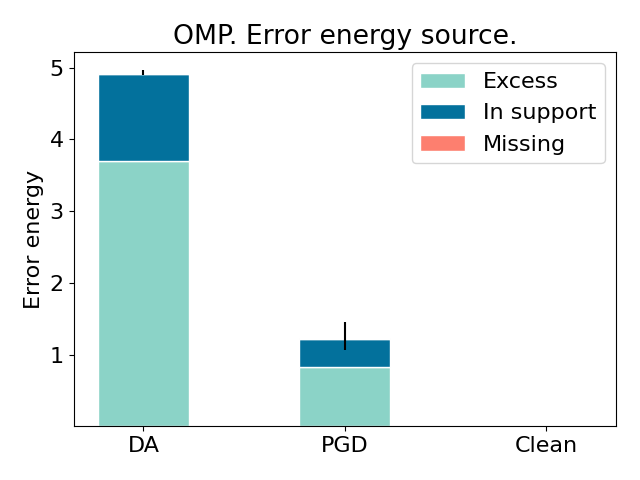}
    \label{fig:da_on_omp}}
    \hfil
    \subfloat[LASSO]{\includegraphics[width=0.3\linewidth]{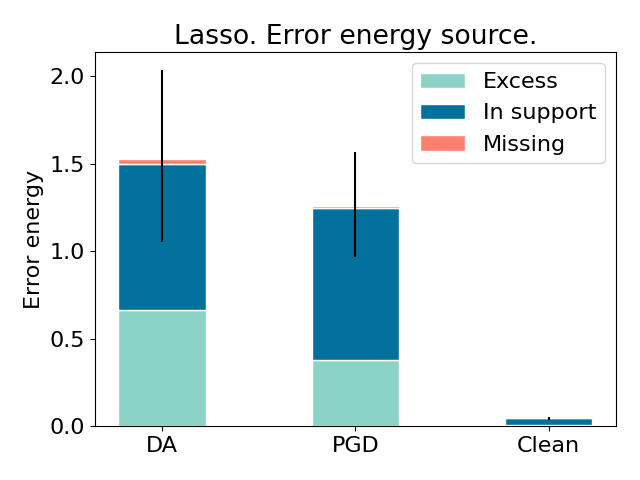}
    \label{fig:da_on_lasso}}
    \hfil
    \subfloat[LISTA]{\includegraphics[width=0.3\linewidth]{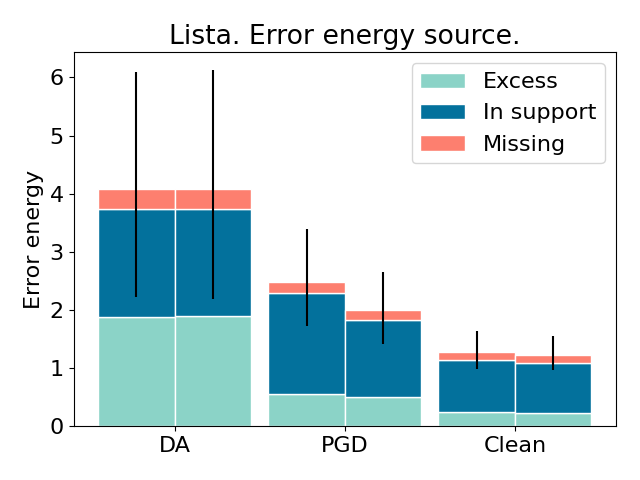}
    \label{fig:da_on_lista}}
\caption{
    Comparison between our dictionary based attack (DA) to PGD on sparse coding algorithms (synthetic data). LISTA: the two columns represent two different trained networks (with the same architecture), where the PGD attack was obtained using the gradients of the left NN. 
    The bars show the mean error of the output sparse codes, with colors referring to the original support (see Section \ref{sec:transf_sparse_coding}), where the black lines show the 0.25 and 0.75 distribution percentiles. DA achieves the maximal mean error, similar in both instances, although it is shared among all examples and agnostic to the networks. PGD is stronger on its originating network (and weaker than our dictionary based attack). DA is stronger than PGD also in OMP and LASSO.} 
\label{fig:DA}
\end{figure*}

\subsection{DICT ATTACK (DA)}
We turn to describe the sparse coding based attack we refer to as "Dict Attack" (DA). Our method optimizes for the highest energy of representation resulting in a bounded input space perturbation, i.e., it solves the optimization problem
\begin{equation}\label{eq:da_optimization}
     \max_{\alpha}   \norm{\alpha}_2 \qquad
     \text{s.t.}\  \norm{D \alpha}_2 \leq \epsilon.
\end{equation}
This can be solved either directly or by the decomposition of the dictionary, where the optimal solution is in the direction of the left eigenvector corresponding to the smallest singular value of $D$; for more details, please see the supplementary material. This problem yields a single attack direction, universal to all data points and coding models, depending only on the dictionary. We now compare the perturbation obtained by our solution, i.e., $D\alpha^\star$, to the PGD attack, on LISTA in the synthetic setting described in Section \ref{sec:transf_sparse_coding}.

The analysis of average squared error in the representation (over a thousand test samples) is presented in Figure \ref{fig:DA} showing the advantage of our dictionary based attack. Notice that the DA attack is universal to all input examples. Moreover, as we are in the synthetic case, the dictionary is given, and therefore it is also agnostic to the LISTA architecture and its trained weights, as opposed to the PGD attack which is designed per network and example (whitebox). 
Notice also that we get similar results for different LISTA networks (represented by the two different bars in Figure~\ref{fig:da_on_lista}), which have different weights. While the DA induced error was similar in both networks, cross PGD attacks were less effective on the opposite network (central two bars in \ref{fig:da_on_lista}). The left bar for PGD with LISTA is for attacking the same network and the right bar is for transferring the attack from one LISTA to another, to which PGD did not have access. While the cross PGD attacks are still effective, there is a decrease in the error of about $20\%$. Yet, the dictionary attacks caused a similar, higher, error. 
Thus, our dictionary based attack has higher transferability than PGD attacks, even though it is oblivious to the networks, in the sparse coding task.
For the completeness of our analysis, we also show the effect of DA on the classical sparse coding algorithms (Figures \ref{fig:da_on_omp}, \ref{fig:da_on_lasso}). Their stopping criterion is defined by a tolerance value equal to the LISTA error on clean input.
Our attack damages these methods as well and is stronger than the adversarial perturbation obtained by PGD on LISTA.

Since we trained LISTA with the $\ell_2$ loss directly to reconstruct the sparse code, a PGD attack can be seen as a practical optimization method for solving a similar problem to \eqref{eq:da_optimization} that only uses the back propagated gradients of the network but without using $D$ explicitly. Thus, no wonder that directly solving \eqref{eq:da_optimization} achieves a better adversarial attack in the task of sparse coding. This motivates the use of a learned forward mapping from the latent space to the input space in the calculations of adversarial attacks; We leave this to future work. We next turn to use this analogy for applying DA to other tasks.

\subsection{BEYOND SPARSE CODING}\label{sec:classification}


The fact that DA performs well in the sparse coding task may not be surprising. Yet, the question is whether we can apply it also to general neural networks for a classification task. This will demonstrate further that adversarial vulnerability is very much related to sparse representations.

General neural networks, trained to perform various tasks such as classification, can be virtually divided into two blocks. First, an encoder block, that encodes the input to some latent representation (often referred to as the 'embedding', 'feature vector', etc.). Then there is a task block, that performs some task on the learned embedding, e.g., classification. In the former case of sparse coding, the encoder block is comprised of the entire LISTA network, and the task block is the identity function. 
In other tasks, however, the encoder block learns a general representation, that is not explicitly related to the dictionary but rather designed by the task, and trained with a different loss. 
We turn to examine whether a dictionary based attack is still applicable in such a case, where the embedding is not explicitly sparse. 

When we come to think about designing an attack for a general task, a natural development would be to solve the task equivalent version of \eqref{eq:da_optimization}, utilizing the dictionary. If we learn a simple linear classifier on the sparse code of the input image (under a dictionary that is not related to the target classification task), we can devise an adversarial perturbation by finding the maximal step in the direction leading to the closest faulty target class. 
Let $W$ be a linear classifier on the sparse code ($W$ has columns as the number of classes and rows as the dimension of the sparse code). The predicted clean class is $\hat{y} = \arg\max \alpha_0^T W$, where $\alpha_0$ is the sparse code of an example data point under the dictionary $D$. In order to fool the classifier, we need to lower the largest component in $\hat{y}$ and enhance the second largest one. Considering a linear classifier and a linear sparse coder, this is equivalent to solving
\begin{equation}\label{eq:da_optimization_cls}
     \max_{\alpha}   \alpha^T (w_{i^\star} - w_c) \qquad
     \text{s.t.}\  \norm{D \alpha}_2 \leq \epsilon,
\end{equation}
where $c$ is the true class and
\begin{equation}
    i^\star = \underset{i \ne c}{\arg\min} \ \alpha_0^T (w_c - w_i),  
\end{equation}
i.e., $i^\star$ is the closest target to the true class. Thus, we find the best perturbation in the code domain, in the direction between the two classes, which is bounded in the input domain. Notice that for calculating the perturbation in the image domain, we simply use $D\alpha^\star$. Although this formulation is limited with respect to the general attack model, i.e. causing a general misclassification, this geometric reasoning results in adversarial perturbations and can be applied also in the general setting as we show hereafter.

Notice that this attack is universal per source and target class pair; many input examples might have the same optimal adversarial perturbation. The dependence of a specific example is only on the choice of the target class. Of course, a more sophisticated classifier such as a neural network might have trickier decision boundaries, where the optimal direction in the features domain is not as clear. Nevertheless, this simple attack model still allows us to find universal, black box perturbations, compromising the accuracy of much more complex models. Thus, we are able to establish the connection discussed in Section \ref{sec:SC_analysis_of_cls_pgd} in the other direction, i.e., attacking the sparse code affects the neural network representation in an adversarial manner, even though they are not related. Moreover, this can be seen as a means for the translation of adversarial perturbations from one task to another, implying some transferability between tasks \cite{transferability_1}.

\begin{figure}
\centering
    \subfloat[ResNet18, MNIST]{\includegraphics[width=0.45\linewidth]{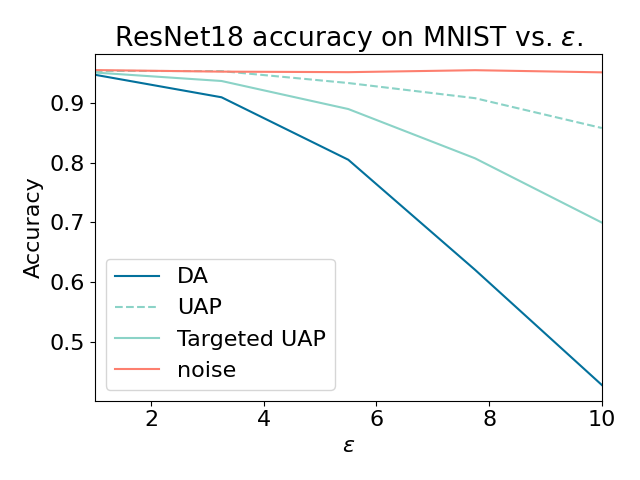}
              \label{fig:da_vs_noise_classification_mnist}}
    \hfil
   \subfloat[WideResNet28, CIFAR10]{\includegraphics[width=0.45\linewidth]{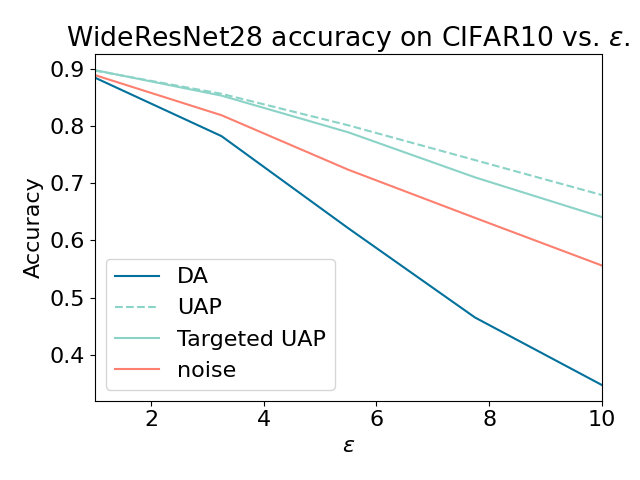}
              \label{fig:da_vs_noise_classification_cifar}}
\caption{
    DA compared to the targeted UAP and random noise, measured in a \textbf{classification} network accuracy, with different attack budgets. Robust accuracy of increasingly stronger $\ell_2$ attacks on WideResNet28 and CIFAR10 (right), ResNet18, and MNIST. The attacks are oblivious to the networks and only use the data dictionaries and linear classifiers. This demonstrates that attacks on the sparse code also hurt the performance of general neural networks, and shows how the sparse coding model can be used to devise universal perturbations that transfer better to different networks.
    } 
\label{fig:da_vs_noise_classification}
\end{figure}

To examine our approach, we start with the following experiment on MNIST. For the task network, we train a generic ResNet18 model, and the sparse coding dictionary is learned using ConvLista \cite{Raja_conv_dict}. Although the encoder of ConvLista is not linear, the solution of \eqref{eq:da_optimization_cls} is a good approximation.
On top of the learned sparse code, we learn a linear classifier $W$, separately. Given the learned dictionary and classifier, we solve \eqref{eq:da_optimization_cls} for every source and target class pair. The optimization process is done once, offline; At inference, we choose the closest target class per example with respect to the output scores of the linear classifier.
We compare our method to the universal adversarial perturbation (UAP) in \cite{universal_adv_attacks} and to its targeted version presented in \cite{targeted_uap}, calculated on the sparse code classifier, using the standard implementations in \cite{art_toolbox}. To be fair, in the targeted case we chose the target class in a similar way to our application of DA, according to the scores of the linear classifier at inference. 
Thus, both methods use only the linear classifier and learned dictionary to generate a set of semi-universal attacks;  \cite{targeted_uap} calculates an attack per target class, while we also consider the original class. The perturbations are then used to attack the task network (ResNet18 for MNIST).
Figure \ref{fig:da_vs_noise_classification_mnist} shows the accuracy of the task network for different values of attack budgets $\epsilon$. DA is more efficient compared to the original and targeted UAP and compared to noise of the same norm. This demonstrates the dual universality of attacks on the sparse representation, both between models and across examples.


We repeat the same experiment using CIFAR-10 and WideResNet28, which is a commonly used model for robustness benchmarking with this dataset (see for example \cite{adversarial_robustness_benchmark, pang_2022_wide_resnet, gowal_2021_wide_resnet}). In this case, we fine-tuned the LISTA convolutional dictionary while learning the linear classifier, to allow better accuracy, while maintaining the visual integrity of the sparse reconstruction. In Figure \ref{fig:da_vs_noise_classification_cifar} we see again the advantage of DA attack over the targeted and original UAP in \cite{targeted_uap} and added random noise of equivalent magnitude, reaffirming the connection of the sparse representation of natural images to neural network sensitivity. The sparse coding model enables us to use the same information (the linear classifier in this case) to create universal attacks that transfer better to other, more complex networks.
The targeted UAP is weaker than equivalent noise in this case, though we believe this is due to the natural robustness of MNIST to noise, i.e., the noise is more effective on CIFAR-10 (UAP is not weaker on CIFAR-10 compared to MNIST). 

\section{CONCLUSION}\label{sec:conclusion}
In this work, we discuss the connection between sparse representation and adversarial attacks on neural networks. We extend their universality to general sparse coding algorithms, which are usually much simpler than common neural models, and may even be linear. In sparse coding and beyond, we show that attacks on the dictionary hurt the performance of neural networks and that an attack on neural networks is visible under the dictionary. This holds true when the network embedding is not explicitly enforced to be sparse, and has no direct connection to the data dictionary, shedding more light on the relationship between neural network performance and the input data representation. Further investigation of how the data sparsity affects the tradeoff between clean accuracy and robustness, and how the sparse code relates to generally learned features are also of interest; We leave this to future work.

Our analysis can provide a compelling explanation for previously observed phenomena regarding adversarial attacks, such as universality across different images, transferability between various networks, and feature denoising and sparsity advantages. Yet to be analyzed are some other properties of robust neural networks that might have a simple explanation using the sparse representation model, such as weights orthogonality and the superiority of adversarial training over all defense methods \cite{adv_training_is_the_best}. In addition, sparsity can be further utilized to suggest defense methods for neural networks. We leave this to future work.

We believe our approach provides a new toolbox that potentially can lead to new directions for understanding and analyzing neural networks. We would like to harness the rich theory developed for sparsity in signal processing, to further develop the theoretical guarantees of adversarial robustness in general neural networks, through the link we show between these two worlds and the transferability of qualities between them. 

\section*{Acknowledgments} This work was supported by the European research council (ERC-StG 757497 PI Giryes) and the Tel Aviv University Center for AI and Data Science (TAD).

\bibliographystyle{IEEEtran}
\bibliography{bibtex}

\section{Supplementary Material}
\subfile{Supplementary}

\end{document}

%% file: supplementary.tex
\receiveddate{XX Month, XXXX}
\reviseddate{XX Month, XXXX}
\accepteddate{XX Month, XXXX}
\publisheddate{XX Month, XXXX}
\currentdate{XX Month, XXXX}

\title{On The Relationship Between Universal Adversarial Attacks And Sparse Representations - Supplementary Material}

\author{DANA WEITZNER AND RAJA GIRYES}
\affil{School of Electrical Engineering, Faculty of Engineering, Tel Aviv University}
\corresp{CORRESPONDING AUTHOR: Dana Weitzner (e-mail: danaweitzner@mail.tau.ac.il).}
\markboth{On The Relationship Between Universal Adversarial Attacks And Sparse Representations}{Weitzner \textit{et al.}}



\maketitle

We now provide additional information regarding the experiments presented in the main paper. 

\textbf{Synthetic sparse coding.}\label{par:synthetic_sparse_coding}
In this experiment, we compared three sparse coding algorithms - OMP, LASSO and LISTA, on synthetic sparse data generated using a Gaussian dictionary $D \in \mathbb{R}^{m \times n}$ with normalized columns. Thus, each data point $x \in \mathbb{R}^{m}$ is sparse under the dictionary $D$, i.e., $x = D \alpha$, where $\alpha \in \mathbb{R}^{n}$ is $s-$sparse. The non zero entries of $\alpha$ are i.i.d standard Gaussian, in locations sampled uniformly at random. Throughout our experiments, $n=500, m=250, s=50$.
We attack LISTA via a standard $\ell_2$ PGD attack, with $\epsilon = 0.3$ and 40 iterations. Each classical algorithm is used in two modes: Promoting more or less sparse solutions. OMP is applied either to match the correct sparsity ($s=50$), or to match the clean error of the LISTA network. LASSO is applied either with $\beta = 10^{-4}, 5 \cdot 10^{-5}$, or with tolerance equal to the clean LISTA error. 
For the classical algorithms we used the Scikit-learn implementation \cite{scikit}. LISTA was implemented in Pytorch \cite{pytorch}, with $K=16$ folds.

\textbf{ConvLista.}
For the non-synthetic datasets, i.e., CIFAR-10 and MNIST, we had to learn a dictionary to have a sparse representation of the data. To this end, we have used ConvLISTA \cite{Raja_conv_dict}, a model that combines the LISTA structure together with a convolutional dictionary. We used a model with $K=3$ ISTA iterations, where the convolutional dictionary is of spatial resolution of $7 \times 7$ and $64$ channels.

\textbf{ConvLista training.}
For the non-synthetic datasets, i.e., CIFAR-10 and MNIST, we had to learn a dictionary to have a sparse representation of the data. We trained a ConvLISTA for 100 epochs using the Adam optimizer and learning rate of $10^{-4}$. We used a combination of SSIM and $\ell_1$ losses, with factors of $0.8, 0.2$, respectively.

\textbf{Classification.}
All of the classification experiments were performed on non-synthetic data. For CIFAR-10, we trained a standard WideResNet28 \cite{wide_resnet} for 100 epochs with cross entropy loss, using SGD with $0.9$ momentum and an initial learning rate of $0.01$ with a $0.1$ decay rate at one and two thirds of the training epochs. For MNIST, we trained ResNet18 \cite{resnet}, where we stacked each input image as three channels, in a similar fashion to CIFAR-10.

\textbf{Adversarial-attack model.} In the sparsity context, the adversarial attack model targets the reconstruction error, which is MSE. Thus, the perturbation is calculated via an optimization process that maximizes the loss such that the perturbation norm is bounded by epsilon. In classification, we use the standard framework where the attack targets the classification loss (cross-entropy) with the standard norm and epsilon values, as used in the common robustness benchmarks (see \cite{adversarial_robustness_benchmark} for example), and are stated below.

\textbf{PGD attacks and equivalent noise.}
For the PGD attacks, we used the CleverHans library Pytorch implementation \cite{cleverhans, pytorch}, with 40 iterations and a step size of 0.01. In the synthetic experiments, we used $\epsilon = 0.5$ in $\ell_2$ experiments and $\epsilon = 0.05$ or $\ell_\infty$ experiments, which have similar corresponding norms in the Gaussian case, i.e., for a Gaussian vector $x$, if $\norm{x}_2 = 0.5$ then $\norm{x}_\infty \approx 0.05$. In the experiments on CIFAR10 and MNIST, we used $\epsilon = 128/255$ in $\ell_2$ experiments and $\epsilon = 8/255$ in $\ell_\infty$ experiments, which are standard in adversarial robustness benchmarks on these datasets (for example, \cite{adversarial_robustness_benchmark}).

\textbf{Dict attack implementation.}
The optimization problem (3) in the main paper can be solved via standard optimization tools (e.g. using the Scipy.optimize library \cite{scipy}) or by using the decomposition of the dictionary in the following manner: Let $D = USV^T$ be the singular value decomposition of $D$ and let $\alpha = Va$. Thus, we have that $\norm{D \alpha}_2^2 = \norm{S a}_2^2 = \sum_i S_{ii} a_i^2 \leq \epsilon^2$ and $\norm{\alpha}_2 = \norm{a}_2$. The optimal solution to $a$ is given by $a^\star = \frac{\epsilon}{S_{min}} v_{min}$, where $S_{min}$ and $v_{min}$ correspond to the smallest singular value and singular vector of $D$ respectively. Finally, we have that the optimal attack is equal to $\epsilon u_{min}$, where $u_{min}$ corresponds to the left eigenvector with the smallest singular value of  $D$. 

\textbf{AutoAttack}
We now briefly describe the attacks included in Auto attack. For more details, please see \cite{AutoAttack}. 
APGD is an automatic PGD in the sense that it is a size-free version of PGD on the cross-entropy loss. 
APGDT is similar only applied to the Difference of Logits Ratio (DLR) loss.
FAB \cite{croce_2020_fab} minimizes the norm of the perturbation necessary to achieve a misclassification.
Square Attack \cite{andriushchenko_2020_square_attack} is a query-efficient score-based attack for norm-bounded perturbations which uses random search and
does not exploit any gradient approximation.

\textbf{Evaluation metrics.} Note that the main purpose of our paper is to show the relationship between sparsity and adversarial attack of neural networks, hence many of our experiments compare the distribution of adversarial perturbations to the distribution of random noise. This kind of comparison results in histograms, or other qualitative plots. We have added  statistical hypothesis tests and confidence intervals to further validate and quantify the empirical results presented in the plots. Other than that, the evaluation metric in the sparse coding experiments is the mean squared error, and test accuracy in the classification settings.


%% file: main.bbl
\begin{thebibliography}{10}
\providecommand{\url}[1]{#1}
\csname url@samestyle\endcsname
\providecommand{\newblock}{\relax}
\providecommand{\bibinfo}[2]{#2}
\providecommand{\BIBentrySTDinterwordspacing}{\spaceskip=0pt\relax}
\providecommand{\BIBentryALTinterwordstretchfactor}{4}
\providecommand{\BIBentryALTinterwordspacing}{\spaceskip=\fontdimen2\font plus
\BIBentryALTinterwordstretchfactor\fontdimen3\font minus
  \fontdimen4\font\relax}
\providecommand{\BIBforeignlanguage}[2]{{%
\expandafter\ifx\csname l@#1\endcsname\relax
\typeout{** WARNING: IEEEtran.bst: No hyphenation pattern has been}%
\typeout{** loaded for the language `#1'. Using the pattern for}%
\typeout{** the default language instead.}%
\else
\language=\csname l@#1\endcsname
\fi
#2}}
\providecommand{\BIBdecl}{\relax}
\BIBdecl

\bibitem{adv_examples_introduction}
C.~Szegedy, W.~Zaremba, I.~Sutskever, J.~Bruna, D.~Erhan, I.~Goodfellow, and
  R.~Fergus, ``Intriguing properties of neural networks,'' 2nd International
  Conference on Learning Representations, ICLR 2014.

\bibitem{defence_1}
N.~Papernot, P.~McDaniel, X.~Wu, S.~Jha, and A.~Swami, ``Distillation as a
  defense to adversarial perturbations against deep neural networks,'' in
  \emph{IEEE symposium on security and privacy (SP)}, 2016, pp. 582--597.

\bibitem{defence_2}
E.~Wong and Z.~Kolter, ``Provable defenses against adversarial examples via the
  convex outer adversarial polytope,'' in \emph{ICML}, 2018.

\bibitem{defence_3}
A.~Madry, A.~Makelov, L.~Schmidt, D.~Tsipras, and A.~Vladu, ``Towards deep
  learning models resistant to adversarial attacks,'' in \emph{International
  Conference on Learning Representations}, 2018.

\bibitem{attack_1}
N.~Carlini and D.~Wagner, ``Towards evaluating the robustness of neural
  networks,'' in \emph{2017 ieee symposium on security and privacy (sp)}.\hskip
  1em plus 0.5em minus 0.4em\relax IEEE, 2017, pp. 39--57.

\bibitem{attack_2}
A.~Kurakin, I.~J. Goodfellow, and S.~Bengio, ``Adversarial machine learning at
  scale,'' in \emph{International Conference on Learning Representations},
  2017.

\bibitem{explaining_1}
I.~J. Goodfellow, J.~Shlens, and C.~Szegedy, ``Explaining and harnessing
  adversarial examples,'' \emph{International Conference on Learning
  Representations}, 2015.

\bibitem{explaining_2}
A.~Fawzi, H.~Fawzi, and O.~Fawzi, ``Adversarial vulnerability for any
  classifier,'' \emph{Advances in neural information processing systems}, 2018.

\bibitem{explaining_3}
A.~Shafahi, W.~R. Huang, C.~Studer, S.~Feizi, and T.~Goldstein, ``Are
  adversarial examples inevitable?'' in \emph{ICLR}, 2019.

\bibitem{explaining_4}
L.~Schmidt, S.~Santurkar, D.~Tsipras, K.~Talwar, and A.~Madry, ``Adversarially
  robust generalization requires more data,'' in \emph{Proceedings of the 32nd
  International Conference on Neural Information Processing Systems}, 2018, pp.
  5019--5031.

\bibitem{explaining_5}
M.~Khoury and D.~Hadfield-Menell, ``On the geometry of adversarial examples,''
  \emph{arXiv preprint arXiv:1811.00525}, 2018.

\bibitem{explaining_6}
S.~Bubeck, Y.~T. Lee, E.~Price, and I.~Razenshteyn, ``Adversarial examples from
  computational constraints,'' in \emph{International Conference on Machine
  Learning}.\hskip 1em plus 0.5em minus 0.4em\relax PMLR, 2019, pp. 831--840.

\bibitem{universal_adv_attacks}
S.-M. Moosavi-Dezfooli, A.~Fawzi, O.~Fawzi, and P.~Frossard, ``Universal
  adversarial perturbations,'' in \emph{Proceedings of the IEEE conference on
  computer vision and pattern recognition}, 2017.

\bibitem{explain_universality_3}
A.~Ilyas, S.~Santurkar, D.~Tsipras, L.~Engstrom, B.~Tran, and A.~Madry,
  ``Adversarial examples are not bugs, they are features,'' \emph{Advances in
  neural information processing systems}, vol.~32, 2019.

\bibitem{explain_universality_4}
K.~R. Mopuri, P.~K. Uppala, and R.~V. Babu, ``Ask, acquire, and attack:
  Data-free {UAP} generation using class impressions,'' in \emph{ECCV}, 2018.

\bibitem{explain_universality_2}
S.~Jetley, N.~Lord, and P.~Torr, ``With friends like these, who needs
  adversaries?'' \emph{Advances in neural information processing systems},
  vol.~31, 2018.

\bibitem{transferability_3}
N.~Papernot, P.~McDaniel, I.~Goodfellow, S.~Jha, Z.~B. Celik, and A.~Swami,
  ``Practical black-box attacks against deep learning systems using adversarial
  examples,'' \emph{Proceedings of the 2017 ACM on Asia Conference on Computer
  and Communications Security}, pp. 506--519, 2017.

\bibitem{transferability_4}
C.~Xie, Z.~Zhang, Y.~Zhou, S.~Bai, J.~Wang, Z.~Ren, and A.~L. Yuille,
  ``Improving transferability of adversarial examples with input diversity,''
  in \emph{Proceedings of the IEEE/CVF Conference on Computer Vision and
  Pattern Recognition (CVPR)}, June 2019.

\bibitem{transferability_6}
Q.~Huang, I.~Katsman, H.~He, Z.~Gu, S.~Belongie, and S.-N. Lim, ``Enhancing
  adversarial example transferability with an intermediate level attack,'' in
  \emph{Proceedings of the IEEE/CVF International Conference on Computer Vision
  (ICCV)}, October 2019.

\bibitem{transferability_1}
N.~Papernot, P.~McDaniel, and I.~Goodfellow, ``Transferability in machine
  learning: from phenomena to black-box attacks using adversarial samples,''
  \emph{arXiv preprint arXiv:1605.07277}, 2016.

\bibitem{transferability_2}
H.~Hosseini, Y.~Chen, S.~Kannan, B.~Zhang, and R.~Poovendran, ``Blocking
  transferability of adversarial examples in black-box learning systems,''
  \emph{arXiv preprint arXiv:1703.04318}, 2017.

\bibitem{transferable}
Y.~Liu, X.~Chen, C.~Liu, and D.~Song, ``Delving into transferable adversarial
  examples and black-box attacks,'' in \emph{ICLR}, 2017.

\bibitem{double_universal}
Y.~Li, S.~Bai, C.~Xie, Z.~Liao, X.~Shen, and A.~Yuille, ``Regional homogeneity:
  Towards learning transferable universal adversarial perturbations against
  defenses,'' in \emph{European Conference on Computer Vision}.\hskip 1em plus
  0.5em minus 0.4em\relax Springer, 2020, pp. 795--813.

\bibitem{SC_linear_classification_robustness}
Y.~Romano, A.~Aberdam, J.~Sulam, and M.~Elad, ``Adversarial noise attacks of
  deep learning architectures: Stability analysis via sparse-modeled signals,''
  \emph{Journal of Mathematical Imaging and Vision}, vol.~62, no.~3, pp.
  313--327, 2020.

\bibitem{sulam}
J.~Sulam, R.~Muthukumar, and R.~Arora, ``Adversarial robustness of supervised
  sparse coding,'' \emph{Advances in Neural Information Processing Systems},
  vol.~33, pp. 2110--2121, 2020.

\bibitem{encoder_gap}
N.~Mehta and A.~Gray, ``Sparsity-based generalization bounds for predictive
  sparse coding,'' in \emph{International Conference on Machine
  Learning}.\hskip 1em plus 0.5em minus 0.4em\relax PMLR, 2013, pp. 36--44.

\bibitem{Sulam_new}
R.~Muthukumar and J.~Sulam, ``Adversarial robustness of sparse local lipschitz
  predictors,'' \emph{arXiv preprint arXiv:2202.13216}, 2022.

\bibitem{activation_pruning}
G.~S. Dhillon, K.~Azizzadenesheli, J.~D. Bernstein, J.~Kossaifi, A.~Khanna,
  Z.~C. Lipton, and A.~Anandkumar, ``Stochastic activation pruning for robust
  adversarial defense,'' in \emph{International Conference on Learning
  Representations}, 2018.

\bibitem{parseval}
M.~Cisse, P.~Bojanowski, E.~Grave, Y.~Dauphin, and N.~Usunier, ``Parseval
  networks: Improving robustness to adversarial examples,'' in
  \emph{Proceedings of the 34th International Conference on Machine Learning},
  ser. Proceedings of Machine Learning Research, 2017.

\bibitem{Lista}
K.~Gregor and Y.~LeCun, ``Learning fast approximations of sparse coding,'' in
  \emph{Proceedings of the 27th international conference on international
  conference on machine learning}, 2010, pp. 399--406.

\bibitem{omp}
J.~A. Tropp and A.~C. Gilbert, ``Signal recovery from random measurements via
  orthogonal matching pursuit,'' \emph{IEEE Transactions on Information
  Theory}, vol.~53, no.~12, pp. 4655--4666, 2007.

\bibitem{adversarial_robustness_benchmark}
F.~Croce, M.~Andriushchenko, V.~Sehwag, E.~Debenedetti, N.~Flammarion,
  M.~Chiang, P.~Mittal, and M.~Hein, ``Robustbench: a standardized adversarial
  robustness benchmark,'' in \emph{Thirty-fifth Conference on Neural
  Information Processing Systems Datasets and Benchmarks Track (Round 2)}.

\bibitem{pang_2022_wide_resnet}
T.~Pang, M.~Lin, X.~Yang, J.~Zhu, and S.~Yan, ``Robustness and accuracy could
  be reconcilable by (proper) definition,'' \emph{ICML}, 2022.

\bibitem{gowal_2021_wide_resnet}
S.~Gowal, S.-A. Rebuffi, O.~Wiles, F.~Stimberg, D.~A. Calian, and T.~A. Mann,
  ``Improving robustness using generated data,'' \emph{Advances in Neural
  Information Processing Systems}, 2021.

\bibitem{AutoAttack}
F.~Croce and M.~Hein, ``Reliable evaluation of adversarial robustness with an
  ensemble of diverse parameter-free attacks,'' in \emph{International
  conference on machine learning}.\hskip 1em plus 0.5em minus 0.4em\relax PMLR,
  2020, pp. 2206--2216.

\bibitem{Raja_conv_dict}
H.~Sreter and R.~Giryes, ``Learned convolutional sparse coding,'' in \emph{2018
  IEEE International Conference on Acoustics, Speech and Signal Processing
  (ICASSP)}.\hskip 1em plus 0.5em minus 0.4em\relax IEEE, 2018, pp. 2191--2195.

\bibitem{tiny_imagenet}
Y.~Le and X.~Yang, ``Tiny imagenet visual recognition challenge,'' \emph{CS
  231N}, vol.~7, no.~7, pp. 3--12, 2015.

\bibitem{targeted_uap}
H.~Hirano and K.~Takemoto, ``Simple iterative method for generating targeted
  universal adversarial perturbations,'' \emph{Algorithms}, vol.~13, no.~11, p.
  268, 2020.

\bibitem{art_toolbox}
M.-I. Nicolae, M.~Sinn, M.~N. Tran, B.~Buesser, A.~Rawat, M.~Wistuba,
  V.~Zantedeschi, N.~Baracaldo, B.~Chen, H.~Ludwig \emph{et~al.}, ``Adversarial
  robustness toolbox v1. 0.0,'' \emph{arXiv preprint arXiv:1807.01069}, 2018.

\bibitem{adv_training_is_the_best}
A.~Athalye, N.~Carlini, and D.~Wagner, ``Obfuscated gradients give a false
  sense of security: Circumventing defenses to adversarial examples,'' in
  \emph{International conference on machine learning}.\hskip 1em plus 0.5em
  minus 0.4em\relax PMLR, 2018, pp. 274--283.

\bibitem{scikit}
F.~Pedregosa, G.~Varoquaux, A.~Gramfort, V.~Michel, B.~Thirion, O.~Grisel,
  M.~Blondel, P.~Prettenhofer, R.~Weiss, V.~Dubourg, J.~Vanderplas, A.~Passos,
  D.~Cournapeau, M.~Brucher, M.~Perrot, and E.~Duchesnay, ``Scikit-learn:
  Machine learning in {P}ython,'' \emph{Journal of Machine Learning Research},
  vol.~12, pp. 2825--2830, 2011.

\bibitem{pytorch}
\BIBentryALTinterwordspacing
A.~Paszke, S.~Gross, F.~Massa, A.~Lerer, J.~Bradbury, G.~Chanan, T.~Killeen,
  Z.~Lin, N.~Gimelshein, L.~Antiga, A.~Desmaison, A.~Kopf, E.~Yang, Z.~DeVito,
  M.~Raison, A.~Tejani, S.~Chilamkurthy, B.~Steiner, L.~Fang, J.~Bai, and
  S.~Chintala, ``Pytorch: An imperative style, high-performance deep learning
  library,'' in \emph{Advances in Neural Information Processing Systems
  32}.\hskip 1em plus 0.5em minus 0.4em\relax Curran Associates, Inc., 2019,
  pp. 8024--8035. [Online]. Available:
  \url{http://papers.neurips.cc/paper/9015-pytorch-an-imperative-style-high-performance-deep-learning-library.pdf}
\BIBentrySTDinterwordspacing

\bibitem{wide_resnet}
S.~Zagoruyko and N.~Komodakis, ``Wide residual networks,'' \emph{British
  Machine Vision Conference}, 2016.

\bibitem{resnet}
K.~He, X.~Zhang, S.~Ren, and J.~Sun, ``Deep residual learning for image
  recognition,'' in \emph{Proceedings of the IEEE conference on computer vision
  and pattern recognition}, 2016, pp. 770--778.

\bibitem{cleverhans}
N.~Papernot, F.~Faghri, N.~Carlini, I.~Goodfellow, R.~Feinman, A.~Kurakin,
  C.~Xie, Y.~Sharma, T.~Brown, A.~Roy, A.~Matyasko, V.~Behzadan,
  K.~Hambardzumyan, Z.~Zhang, Y.-L. Juang, Z.~Li, R.~Sheatsley, A.~Garg,
  J.~Uesato, W.~Gierke, Y.~Dong, D.~Berthelot, P.~Hendricks, J.~Rauber, and
  R.~Long, ``Technical report on the cleverhans v2.1.0 adversarial examples
  library,'' \emph{arXiv preprint arXiv:1610.00768}, 2018.

\bibitem{scipy}
P.~Virtanen, R.~Gommers, T.~E. Oliphant, M.~Haberland, T.~Reddy, D.~Cournapeau,
  E.~Burovski, P.~Peterson, W.~Weckesser, J.~Bright, S.~J. {van der Walt},
  M.~Brett, J.~Wilson, K.~J. Millman, N.~Mayorov, A.~R.~J. Nelson, E.~Jones,
  R.~Kern, E.~Larson, C.~J. Carey, {\.I}.~Polat, Y.~Feng, E.~W. Moore,
  J.~{VanderPlas}, D.~Laxalde, J.~Perktold, R.~Cimrman, I.~Henriksen, E.~A.
  Quintero, C.~R. Harris, A.~M. Archibald, A.~H. Ribeiro, F.~Pedregosa, P.~{van
  Mulbregt}, and {SciPy 1.0 Contributors}, ``{{SciPy} 1.0: Fundamental
  Algorithms for Scientific Computing in Python},'' \emph{Nature Methods},
  vol.~17, pp. 261--272, 2020.

\bibitem{croce_2020_fab}
F.~Croce and M.~Hein, ``Minimally distorted adversarial examples with a fast
  adaptive boundary attack,'' in \emph{International Conference on Machine
  Learning}.\hskip 1em plus 0.5em minus 0.4em\relax PMLR, 2020, pp. 2196--2205.

\bibitem{andriushchenko_2020_square_attack}
M.~Andriushchenko, F.~Croce, N.~Flammarion, and M.~Hein, ``Square attack: a
  query-efficient black-box adversarial attack via random search,'' in
  \emph{European Conference on Computer Vision}.\hskip 1em plus 0.5em minus
  0.4em\relax Springer, 2020, pp. 484--501.

\end{thebibliography}
